\documentclass{vgtc}                          

\ifpdf
  \pdfoutput=1\relax                   
  \usepackage{graphicx}                
  \DeclareGraphicsExtensions{.pdf,.png,.jpg,.jpeg} 
\else
  \usepackage{graphicx}                
  \DeclareGraphicsExtensions{.eps}     
\fi%

\graphicspath{{figures/}{pictures/}{images/}{./}} 

\usepackage{microtype}                 
\PassOptionsToPackage{warn}{textcomp}  
\usepackage{textcomp}                  
\usepackage{mathptmx}                  
\usepackage{times}                     
\usepackage{cite}                      
\usepackage{tabu}                      
\usepackage{booktabs}                  

\usepackage{xspace}
\usepackage{enumitem}
\usepackage{multirow}
\usepackage[table]{xcolor}
\usepackage{listings}
\usepackage{amssymb}

\definecolor{codegreen}{rgb}{0,0.6,0}
\definecolor{codegray}{rgb}{0.5,0.5,0.5}
\definecolor{codepurple}{rgb}{0.58,0,0.82}
\definecolor{backcolour}{rgb}{0.95,0.95,0.95}

\lstdefinestyle{mystyle}{
    backgroundcolor=\color{backcolour},   
    commentstyle=\color{codegray},
    keywordstyle=\color{magenta},
    numberstyle=\tiny\color{codepurple},
    stringstyle=\color{codegreen},
    basicstyle=\ttfamily\footnotesize,
    breakatwhitespace=false,         
    breaklines=true,                 
    captionpos=b,                    
    keepspaces=true,                 
    numbers=left,                    
    numbersep=5pt,                  
    showspaces=false,                
    showstringspaces=false,
    showtabs=false,                  
    tabsize=1
}

\def\library{ZADU\xspace}
\def\vislib{ZADUVis\xspace}

\onlineid{1036}

\vgtccategory{Research}

\vgtcinsertpkg

\title{\textit{\library}: A Python Library for Evaluating the Reliability of \\ Dimensionality Reduction Embeddings}

\author{
    Hyeon Jeon\thanks{e-mail: \{hj, archo, jhjang, shlee\}@hcil.snu.ac.kr} { }$^1$, %
    Aeri Cho$^*${ }$^1$,
    Jinhwa Jang$^*${ }$^1${ }$^3$,
    Soohyun Lee$^*${ }$^1$,
    Jake Hyun$^\S${ }$^1$, \\
    Hyung-Kwon Ko\thanks{hyungkwonko@gmail.com} { }$^2$,
    Jaemin Jo\thanks{jmjo@skku.edu} { }$^4$, and
    Jinwook Seo\thanks{\{jakehyun, jseo\}@snu.ac.kr} { }$^1$ 
    }
\affiliation{\scriptsize $^1$Seoul National University \quad $^2$KAIST \quad $^3$Samsung Electronics \quad $^4$Sungkyunkwan University}

\abstract{
Dimensionality reduction (DR) techniques inherently distort the original structure of input high-dimensional data, producing imperfect low-dimensional embeddings. Diverse distortion measures have thus been proposed to evaluate the reliability of DR embeddings.  
However, implementing and executing distortion measures in practice has so far been time-consuming and tedious. 
To address this issue, we present \library, a Python library that provides distortion measures.
\library is not only easy to install and execute but also
enables comprehensive evaluation of DR embeddings through three key features. First, the library covers a wide range of distortion measures. Second, it automatically optimizes the execution of distortion measures, substantially reducing the running time required to execute multiple measures.
Last, the library informs how individual points contribute to the overall distortions, facilitating the detailed analysis of DR embeddings. By simulating a real-world scenario of optimizing DR embeddings, we verify that our optimization scheme substantially reduces the time required to execute distortion measures. 
Finally, as an application of \library, we present another library called \vislib that allows users to easily create distortion visualizations that depict the extent to which each region of an embedding suffers from distortions.

} 

\CCScatlist{
  \CCScatTwelve{Human-centered computing}{Visu\-al\-iza\-tion}{Visualization design and evaluation methods}{}
}

\begin{document}

\def\sectionautorefname{Section}
\def\subsectionautorefname{Section}
\def\subsubsectionautorefname{Section}

\firstsection{Introduction}
\maketitle

Dimensionality reduction (DR) suffers from inaccuracy.
Although DR is a useful technique for visually analyzing high-dimensional data \cite{nonato19tvcg}, distortion inevitably occurs while moving data from a broad high-dimensional space to a narrow low-dimensional space \cite{nonato19tvcg, martins14cg, jeon21tvcg, jeon22arxiv}. 
Such distortions lower the credibility of data analysis with DR embeddings. 
To avoid such risks of misinterpretation, we need to assess the reliability of the embeddings prior to their usage. 
For this purpose, various distortion measures (e.g., Trustworthiness \& Continuity \cite{lee07springer} and Steadiness \& Cohesiveness \cite{jeon21tvcg}) have been proposed \cite{nonato19tvcg}. 

However, there is a lack of an easy-to-use library that provides distortion measures, which leads to the consumption of researchers' valuable time.
A few research works provide the source code of distortion measures \cite{jeon22vis, ingram15neurocomputing, fujiwara23pvis} (\autoref{tab:measures}). However, researchers need considerable time to install and execute such code. 
For example, they need to manually configure the environment settings and install the dependencies.
Researchers thus often implement distortion measures on their own, but the laboriousness of the task persists.

Given this background, we present \library, a unified and accessible Python library serving distortion measures. 
To save the time needed to install and execute the library, we make \library easily downloadable via the Python package index \texttt{PyPI}.
Moreover, in line with the current trend in DR research \cite{jeon22vis, fujiwara23pvis, jeon21tvcg, pedregosa11jmlr, mcinnes2020arxiv}, \library is compatible with  existing Python machine learning and visualization toolboxes (e.g., \texttt{scikit-learn} \cite{pedregosa11jmlr} and \texttt{matplotlib} \cite{hunter07cse}).

\library differentiates from previous implementations of distortion measures from three perspectives.
First, the library covers a broad range of distortion measures, with a total of 17 provided. This is over three times more than the earlier implementations with the most measures available \cite{jeon22vis}. Hence, researchers do not need to spend time searching for available codes or implementing the codes by themselves.
Second, \library automatically optimizes the execution of multiple measures, substantially reducing the amount of computation time needed. 
Last, \library supports the computation of local pointwise distortions, which illustrates the contribution of each data point to the overall distortions.
By explaining distortions in a fine-grained manner, local distortions enable a more detailed analysis of DR embeddings.

We simulate a real-world scenario of evaluating DR embeddings to assess the extent to which \library optimizes the execution of multiple measures.
The simulation verifies that our optimization substantially reduces the total running time required for executing distortion measures. 
We also demonstrate using \library to create distortion visualizations that depict how and where the embedding suffers from distortions. We have packaged our implementation of distortion visualizations as a library called \vislib, enabling users to readily create the visualizations.

\definecolor{lightred}{RGB}{247, 163, 180}
\definecolor{lightlightred}{RGB}{245, 208, 216}

\newcommand{\lred}{\cellcolor{lightred}}
\newcommand{\llred}{\cellcolor{lightlightred}}

\begin{table*}[h]
  
	\centering%
 \scalebox{0.81}{
  \begin{tabular}{%
	rrcc|c *{14}{c}%
	}
  \toprule
   Type & Measure & Ref.  & \rotatebox{90}{provide pointwise}  \rotatebox{90}{distortions}  & \rotatebox{90}{dreval \cite{soneson22r}} & \rotatebox{90}{McInnes et al. \cite{mcinnes2020arxiv}} &\rotatebox{90}{Ingram et al. \cite{ingram15neurocomputing}} &   \rotatebox{90}{Jeon et al. \cite{jeon21tvcg}} &   \rotatebox{90}{Fujiwara et al. \cite{fujiwara23pvis}} & \rotatebox{90}{Espadoto et al. \cite{espadoto21tvcg}} & \rotatebox{90}{Colange et al. \cite{colange20neurips}} &   \rotatebox{90}{coranking \cite{kraemer18rj}} & \rotatebox{90}{pyclustering \cite{novikov19oss}} &  \rotatebox{90}{scikit-learn \cite{pedregosa11jmlr}}  &   \rotatebox{90}{scipy \cite{virtanen20nature}}  &   \rotatebox{90}{Moor et al. \cite{moor20icml}} &   \rotatebox{90}{Jeon et al. \cite{jeon22vis}}  & \rotatebox{90}{\textbf{ZADU} (Ours)}   \\
  \hline
  \multirow{7}{*}{Local} & Trustworthiness \& Continuity &\cite{venna06nn}  & $\vee$ & \llred \footnotesize{$\bigtriangleup$} & \llred \footnotesize{$\bigtriangleup$} & & & & & \lred \footnotesize{$\bigcirc$} & \lred \footnotesize{$\bigcirc$} & &\llred \footnotesize{$\bigtriangleup$} & & \lred \footnotesize{$\bigcirc$}& \lred \footnotesize{$\bigcirc$}  & \lred \footnotesize{$\bigcirc$} \\
                         & Mean Relative Rank Errors &\cite{lee09neurocomputing}  & $\vee$ & & & & & & & & & & &  & \lred \footnotesize{$\bigcirc$} & \lred \footnotesize{$\bigcirc$}  & \lred \footnotesize{$\bigcirc$} \\   
                         & Local Continuity Meta-Criteria &\cite{chen09jasa}  & $\vee$ & & & & & & & & \lred \footnotesize{$\bigcirc$} & & & & & & \lred \footnotesize{$\bigcirc$} \\   
                         & Neighborhood Hit& \cite{paulovich08tvcg}   & $\vee$ & & & \lred \footnotesize{$\bigcirc$} & & & & & & & & & &  & \lred \footnotesize{$\bigcirc$}\\   
                         & Neighbor Dissimilarity& \cite{fujiwara23pvis}  & & & & & & \lred \footnotesize{$\bigcirc$} & & & & & & & & & \lred \footnotesize{$\bigcirc$} \\   
                         & Class-Aware Trustworthiness \& Continuity &\cite{colange20neurips}  & $\vee$ & & & & & & &\lred \footnotesize{$\bigcirc$}  & & & & & &  & \lred \footnotesize{$\bigcirc$}\\   
                         & Procrustes Measure &\cite{goldberg2009local}  & & & & & & & & & & & & & & & \lred \footnotesize{$\bigcirc$} \\   
  \hline
  \multirow{4}{*}{Cluster-level} & Steadiness \& Cohesiveness& \cite{jeon21tvcg} & $\vee$& & & & \lred \footnotesize{$\bigcirc$} &\lred \footnotesize{$\bigcirc$} & & & & & & & & & \lred \footnotesize{$\bigcirc$}\\
                                 & Distance Consistency& \cite{sips09cgf} & & & & & & & & & & & & & & & \lred \footnotesize{$\bigcirc$} \\
                                 & Internal Clustering Validation Measures &\cite{joia11tvcg}&  & &&  & & & & & & \lred \footnotesize{$\bigcirc$} & \lred \footnotesize{$\bigcirc$}& & & & \lred \footnotesize{$\bigcirc$} \\
                                 & Clustering + External Clustering Validation Measures & \cite{xiang21fig} & & &&  & & & & & & \lred \footnotesize{$\bigcirc$} &\lred \footnotesize{$\bigcirc$} & & & & \lred \footnotesize{$\bigcirc$} \\
  \hline
  \multirow{6}{*}{Global}  & Stress &\cite{kruskal64psycho, kruskal1964nonmetric}  & & & & & & & \lred \footnotesize{$\bigcirc$} & & & & & & \lred \footnotesize{$\bigcirc$} & \lred \footnotesize{$\bigcirc$} & \lred \footnotesize{$\bigcirc$}\\
                          &  Kullback-Leibler Divergence &\cite{hinton02nips} & & & & & & & & & & & & &\lred \footnotesize{$\bigcirc$} &\lred \footnotesize{$\bigcirc$} & \lred \footnotesize{$\bigcirc$} \\  
                          & Distance-to-Measure &\cite{chazal11fcm} & & & & & & & & & & & & & & \lred \footnotesize{$\bigcirc$} & \lred \footnotesize{$\bigcirc$} \\
                          & Topographic Product & \cite{bauer1992quantifying}  & & & & & & & & & & & & & & &\lred \footnotesize{$\bigcirc$} \\
                          & Pearson's correlation coefficient $r$ &\cite{geng2005supervised} &  & & & & & & & & & & & \lred \footnotesize{$\bigcirc$} & & & \lred \footnotesize{$\bigcirc$}\\
                          & Spearman's rank correlation coefficient $\rho$ & \cite{sidney1957nonparametric}  & & & & & & & & & & & & \lred \footnotesize{$\bigcirc$} & & &\lred \footnotesize{$\bigcirc$}  \\

  \bottomrule
  \end{tabular}%
}
\vspace{2mm}
    \caption{The overview of the distortion measures provided by \library (row) and their publicly available implementations (column). The publicized implementations that fully implement the corresponding measures are highlighted in red background and circle. The ones that implement only half of the pair of measures are highlighted in light red background and triangle. \vspace{-4mm}}
  \label{tab:measures}
\end{table*}

\section{Background and Related Work}

\label{sec:relwork}

We discuss the literature associated with distortion measures. We then review the publicly available implementations of the measures.

\subsection{Distortion Measures}

\label{sec:relmeasure}

Distortion measures are functions that accept a high-dimensional data $\mathbf{X} = \{x_i \in \mathbb{R}^D \mid 1 \leq i \leq N \}$ and its low-dimensional embedding $\mathbf{Y} = \{y_i \in \mathbb{R}^d \mid 1 \leq i \leq N \}$ ($d < D$) as input, and then return a score that represents how well the structure of $\mathbf{Y}$ matches that of $\mathbf{X}$. The measures are either developed as a loss function of a DR technique \cite{hinton02nips, colange20neurips} or developed originally, independent of any technique \cite{jeon21tvcg, lee07springer}. 

Distortion measures can be broadly divided into three categories---local measures, global measures, and cluster-level measures---based on their target structural granularity \cite{jeon21tvcg}. 
Local measures evaluate the extent to which the neighborhood structure of $\mathbf{X}$ is preserved in $\mathbf{Y}$. For example, Trustworthiness \& Continuity (T\&C) \cite{venna06nn} and  Mean relative rank error (MRRE) \cite{lee09neurocomputing} assess the degree to which the $k$-nearest neighbors ($k$NN) of each point in $\mathbf{X}$ are no longer neighbors in the $\mathbf{Y}$, and vice versa. Neighborhood Dissimilarity \cite{fujiwara23pvis} measures the level to which the Shared-Nearest Neighbor \cite{ertoz02siam} graph structure is different in $\mathbf{X}$ and $\mathbf{Y}$. 
Next, cluster-level measures evaluate how well the cluster structures of $\mathbf{X}$ are preserved in $\mathbf{Y}$. The cluster is given by clustering algorithms \cite{jeon21tvcg} or class labels \cite{ joia11tvcg}.
Finally, global measures evaluate the extent to which point-pairwise distances remain consistent. For instance, Pearson's correlation coefficient $r$ quantifies how the ranking of point pairs based on their distances varies between $\mathbf{X}$ and $\mathbf{Y}$.

As diverse DR techniques emphasize different facets of data, employing multiple distortion metrics with varying granularity levels is crucial for the comprehensive evaluation of DR embeddings. Therefore, while designing \library, we try not only to maximize the number of supported distortion measures but also to have an even distribution of all types of measures (\autoref{tab:measures}, \autoref{sec:interface}). 




\subsection{Implementations of Distortion Measures}

Despite the importance of reliability evaluations when utilizing DR, there is a lack of a unified implementation that provides distortion measures. The majority of implementations is in publicly accessible repositories contributed by the studies on DR \cite{mcinnes2020arxiv, fujiwara23pvis, cockburn09cs, moor20icml, jeon22vis}. However, each implementation has a limited number of supported distortion measures (\autoref{tab:measures}). Moreover, installing, compiling, and executing from such scattered code is time-consuming.

An alternative way is to use the distortion measures provided by popular machine learning libraries (e.g., \texttt{scikit-learn} \cite{pedregosa11jmlr}). These libraries are easy to install and execute, and also likely to be highly optimized. However, as general-purpose machine learning toolboxes, they offer limited support for distortion measures (\autoref{tab:measures}). We aim to develop a library that (1) is easily downloadable and executable, similar to the widely-used machine learning libraries, while (2) supporting a broader range of distortion measures.

\section{\library}

We first present the supported measures and the interface of \library. We then delve into the functionalities offered by the library that facilitate the efficient and reliable analysis of DR embeddings. 


\subsection{Supported Distortion Measures}

\label{sec:support}

The list of distortion measures to be included in the library is determined through a literature review on DR and their evaluation (\autoref{sec:relwork}). 
Different distortion measures evaluate the preservation of the data structure at varying levels of granularity (e.g., neighborhood, cluster, and global structure; \autoref{sec:relmeasure}). The simultaneous use of multiple measures having different granularity is essential for comprehensively evaluating DR embeddings \cite{jeon22vis, moor20icml, espadoto21tvcg}. 
Thus, we try to maximize both the number of supported measures and the diversity in terms of the structural granularity that the measures focus on. 
As a result, we select seven local measures, four cluster-level measures, and six global measures (\autoref{tab:measures}). Please refer to Appendix A for the detailed procedure for computing each measure.

\subsection{Interface}

\label{sec:interface}

\library provides two different interfaces for executing distortion measures. The first is to use the main class that is named after our library (i.e., \texttt{zadu}).
In designing the main interface, our focus is on reusing both the code and the computing resources so that users can save time.
With regard to reusing code, we force users to write a specification that defines the measures to be executed (\texttt{"id"} in Code 1) along with their hyperparameters (\texttt{"params"}). By reusing the specifications, users can perform an identical evaluation on multiple datasets. This is commonly done in practice to enhance the generalizability of the evaluation \cite{jeon22vis, espadoto21tvcg, moor20icml}.
As for reusing the computing results, we require users to register the original high-dimensional dataset (\texttt{hd}) once, along with its specifications. This dataset can then be reused repeatedly.
This is because the evaluation of DR is usually done by comparing multiple embeddings of a single high-dimensional dataset. 
The distortion measures can then be executed by invoking \texttt{measure} method while giving the embedding (\texttt{ld}) as an argument, which returns the scores from the distortion measures. 

\begin{lstlisting}[
    language=Python, 
    caption={Using the main class of \library  to compute the Trustworthiness \& Continuity (\texttt{tnc}) and Steadiness \& Cohesiveness (\texttt{snc}) scores of a given embedding (\texttt{ld}) based on its original data (\texttt{hd}). },
    label={code:basic}
]
from zadu import zadu

hd, ld = load_datasets()
spec = [{
    "id"    : "tnc",
    "params": { "k": 20 },
}, {
    "id"    : "snc",
    "params": { "k": 30, "clustering": "hdbscan" }
}]

scores = zadu.ZADU(spec, hd).measure(ld)
print("T&C:", scores[0])
print("S&C:", scores[1])

\end{lstlisting}

An alternative interface is to directly invoke the functions that define each distortion measure (\autoref{code:individual}).
However, executing multiple measures in this way does not take advantage of optimization (\autoref{sec:optimize}). Hence, more computation time is needed compared to using the main class (\autoref{code:basic}).

\begin{lstlisting}[
    language=Python, 
    caption={Accessing the internal functions of \library to execute Mean Relative Rank Errors and Pearson's correlation coefficient $r$.},
    label={code:individual}
]
from zadu.measures import *

mrre = mean_relative_rank_error.measure(hd, ld)
pr  = pearson_r.measure(hd, ld)
\end{lstlisting}

\subsection{Functionalities}

We outline the functionalities of \library that enable the effective evaluation and analysis of DR embeddings.

\subsubsection{Optimizing the Execution of Multiple Measures}

\label{sec:optimize}

Utilizing multiple distortion measures simultaneously is common in practice \cite{jeon22vis, moor20icml}. For example, Espadoto et al. \cite{espadoto21tvcg} proposed to aggregate multiple measures by averaging them. 
However, using more measures leads to increased computational demands.

To reduce the computation time running multiple distortion measures, \library automatically optimizes the execution of the measures. 
The primary goal of the optimization is to minimize the computational overhead associated with three key preprocessing blocks: pairwise distance computation, pointwise distance ranking computation, and $k$NN identification.
The pairwise distance computation is done by constructing a distance matrix in both the original and the embedded spaces utilizing a specified distance function (e.g., Euclidean distance or cosine similarity). 
During the pointwise distance ranking computation stage, the ranking of all data points with respect to each individual data point $x$ is set based on their distance from $x$. This is also done in both the original and the embedded spaces. Lastly, $k$NN identification involves locating the top-$k$ closest data points of each point in the original and embedded spaces.

The optimization works as follows. Given a specification (refer to \autoref{sec:interface}), \library extracts a list of requisite preprocessing units. The library then establishes an execution order for the blocks while maximizing the reuse of computed results. For instance, if both the distance matrix and the $k$NN index are needed, the outcome of the former computation is reused to compute the latter. Similarly, if the specifications require the computation of both $k_1$NN and $k_2$NN, where $k_1 > k_2$, the $k_2$NN can be acquired by slicing the $k_1$NN.
Once the execution order and dependencies are ascertained, \library runs preprocessing. The preprocessing results are stored in the RAM and subsequently injected into each function that defines a distortion measure to derive the final scores.

The effectiveness of our optimization increases as more distortion measures are executed simultaneously. We validate that the optimization substantially reduces the execution time of distortion measures through our quantitative evaluation (\autoref{sec:runtime}).

\subsubsection{Computing Pointwise Local Distortions}

\label{sec:pointwise}

\library enables users to obtain local pointwise distortions, which indicate how each point contributes to the overall distortions. 
Such functionality improves the usability of our library as local distortions help users in performing enhanced analysis of DR embeddings. For example, we can aggregate local distortions in class labels to reveal which class is vulnerable to the distortions.
Moreover, we can visualize local distortions \cite{lespinats11cgf, 
jeon21tvcg}, which facilitates a more accurate analysis of the original high-dimensional data \cite{jeon21tvcg}. We discuss this application in more detail in \autoref{sec:app}.

We can obtain local pointwise distortions by raising the \texttt{return\_local} flag. When the flag is raised, the library returns the local distortions along with the aggregated scores (See \autoref{code:distortion}).


\begin{lstlisting}[
    language=Python, 
    caption={Obtaining local pointwise distortion from \library by raising the \texttt{return\_local} flag. If a specified distortion measure produces local pointwise distortion as an intermediate result, it returns a list of pointwise distortions when the flag is raised. },
    label={code:distortion}
]
from zadu import zadu

spec = [{
    "id"    : "dtm",
    "params": {}
}, {
    "id"    : "mrre",
    "params": { "k": 30 }
}]

zadu_obj = zadu.ZADU(spec, hd, return_local=True)
global_, local_ = zadu_obj.measure(ld)
print("MRRE local distortions:", local_[1])

\end{lstlisting}

The computation of pointwise local distortions is available only for some local measures and cluster-level measures (See ``provide pointwise distortions'' column in \autoref{tab:measures}). For example, T\&C and MRREs produce final scores as an average of local distortions. Steadiness \& Cohesiveness \cite{jeon21tvcg} computes pointwise distortion by aggregating partial cluster-level distortions. When the flag is raised, \library returns a list consisting of local pointwise distortions for the available measures; it otherwise returns \texttt{None}. 


\subsection{Implementation}

\library is a Python library that can be installed via \texttt{PyPI} with just a single command. 
Scalability is a key consideration in implementing \library. We maximize the utilization of matrix computation and incorporate highly optimized open-source libraries for computationally heavy tasks (e.g., \texttt{faiss} \cite{johnson19tbd} for $k$NN identification). To simplify the installation and execution, the library runs only on CPUs. 

While implementing the measures, we reuse the previous open-source implementations if available. 
For example, for T\&C, MRRE, Stress, DTM, and KL divergence, we adopt the code provided by Jeon et al. \cite{jeon22vis} (the second last column of \autoref{tab:measures}). For Steadiness \& Cohesiveness, we use the code shared by the authors. We still revise these codes to fit our optimization pipeline (\autoref{sec:optimize}), to make them return local pointwise distortions (\autoref{sec:pointwise}), and to eliminate GPU dependencies. The remaining measures are carefully implemented by referring to the papers in which they were first introduced. The source code is available at \href{https://github.com/hj-n/zadu}{github.com/hj-n/zadu}.

\begin{figure*}
    \centering
    \includegraphics[width=\textwidth]{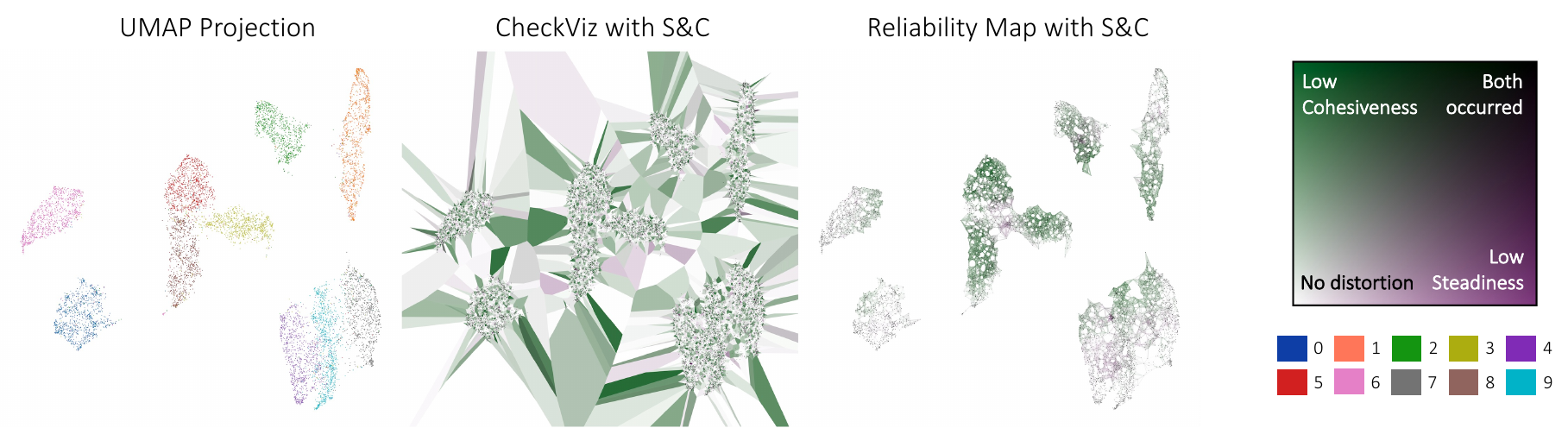}
    \vspace{-6mm}
    \caption{
    The UMAP embedding of the MNIST dataset (leftmost column), and two distortion visualizations generated by \vislib: CheckViz \cite{lespinats11cgf} and the Reliability Map \cite{jeon21tvcg}. 
    The distortion visualizations depict how each region of the given embedding suffers from the distortions that lower the Steadiness \& Cohesiveness (S\&C) scores. The visualizations follow the 2D colormap proposed by Lespinats and Aupetit \cite{lespinats11cgf} (rightmost column).
    Combined with \library, \vislib helps practitioners easily generate distortion visualizations on a \texttt{matplotlib} canvas. \vspace{-4mm}}
    \label{fig:distvis}
\end{figure*}

\section{Runtime Analysis}

\label{sec:runtime}

\subsection{Objectives and Design}

We test whether our optimization pipeline (\autoref{sec:optimize}) reduces the time needed to evaluate DR embeddings.
We simulate a scenario in which we try to optimize the hyperparameters of a DR technique using multiple distortion measures that have common preprocessing blocks.
We evaluate the running time for optimization with and without the optimization. 
We use datasets with diverse characteristics, e.g., the number of points and dimensionality. We compare how the running time of evaluation differs on average as we switch on the optimization. 


\noindent
\textbf{Optimization}
For a given dataset, we measure the time required to run Bayesian optimization \cite{snoek12nips} for finding the optimal value of two hyperparameters in UMAP \cite{mcinnes2020arxiv}: \texttt{nearest neighbors} and \texttt{minimum distance} \cite{mcinnes2020arxiv}. The search range of two hyperparameters is set as (2, 200) and (0.01, 0.99), respectively, following the recommendation of the official documentation\footnote{\href{https://umap-learn.readthedocs.io/en/latest/index.html}{umap-learn.readthedocs.io}}.
For Bayesian optimization, we use the Python implementation of Nogueira \cite{nogueira14github} with the default hyperparameter setting.

\noindent
\textbf{Distortion measures}
For the distortion measures, we use T\&C, MRRE, Steadiness \& Cohesiveness, Distance-to-Measure, and Kullback-Leibler divergence. All the measures share pairwise distance matrix computation as a common preprocessing block. The first three measures also share $k$NN identification. 
As a loss function, we use an average of five measures, following Espadoto et al. \cite{espadoto21tvcg}.

\noindent
\textbf{Datasets}
We apply the optimization to the 96 publicly available high-dimensional datasets gathered by a previous study \cite{jeon22arxiv2}. Every dataset is standardized before applying the optimization process.

\subsubsection{Results}

\begin{figure}
    \centering
    \includegraphics[width=\linewidth]{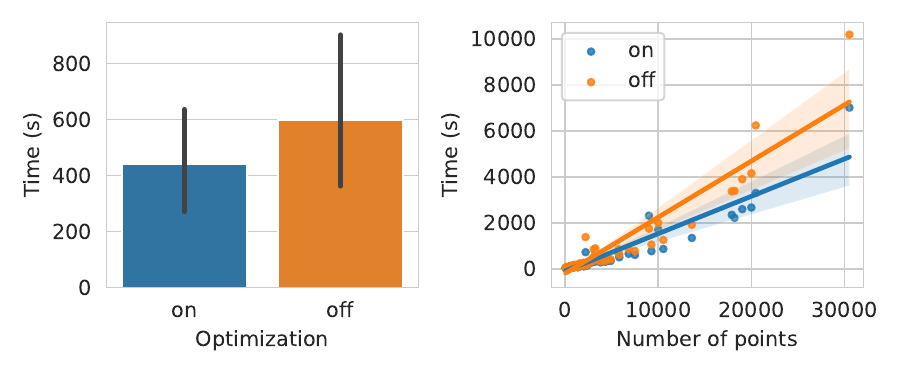}
    \vspace{-7mm}
    \caption{    
    The results of the runtime analysis (\autoref{sec:runtime}). (left) \library{}'s optimization substantially reduces the runtime for the optimization of DR embeddings. (right) The extent to which the optimization reduces the runtime increases as the size of the datasets increases. The shaded area in the right figure depicts the 95\% confidence interval. \vspace{-4mm}}
    \label{fig:runtime}
\end{figure}

\autoref{fig:runtime} depicts the result. \library is 1.5 times faster with optimization than without it on average, verifying the effectiveness of the optimization pipeline. We also discover that the difference in runtime between with and without optimization increases as the number of points in the dataset increases (as indicated by the steeper orange regression line in \autoref{fig:runtime}b compared to the blue regression line). This finding further supports the scalability benefits of \library. Overall, our results demonstrate that \library substantially reduces the time required for practitioners to evaluate DR embeddings.





\subsection{Application: Visualizing Local Distortions}

\label{sec:app}

Various distortion visualization methods \cite{lespinats11cgf, jeon21tvcg} have been proposed to provide insights into the extent to which each region is affected by distortions.
CheckViz \cite{lespinats11cgf} (\autoref{fig:distvis} second column), for example, decomposes the scatterplot that represents a DR embedding using a Voronoi diagram, and then encodes the distortion of each point as a color of the corresponding Voronoi cell. Reliability Map \cite{jeon21tvcg} (\autoref{fig:distvis} third column) constructs an  $k$NN graph in the embedded space and encodes the distortions of each point on the incident graph edges.

We present the implementation of local distortion visualizations as an application of \library. We develop \vislib, a Python library that provides CheckViz and the Reliability Map as representative distortion visualizations. 
\vislib takes local pointwise distortions generated by \library as input and uses them to generate distortion visualizations. 
Integrated with \texttt{matplotlib} \cite{hunter07cse}, \vislib allows users to render a distortion visualization without time-consuming extra implementation (\autoref{code:distvis}). Extending our application to a more complex visual analytics system would be an interesting direction.

\begin{lstlisting}[
    language=Python, 
    caption={Visualizing CheckViz \cite{lespinats11cgf} and the Reliability Map \cite{jeon21tvcg} using \vislib and \texttt{matplotlib}. \vislib gets the embedding and the local distortions made by \library as arguments and generates distortion visualization. The rendered image of this code is depicted in the second (CheckViz) and third (Reliability Map) columns of \autoref{fig:distvis}. },
    label={code:distvis}
]
from zadu import zadu
from zaduvis import zaduvis
import matplotlib.pyplot as plt
from sklearn.manifold import TSNE

## load datasets and generate an embedding
hd = load_mnist()
ld = TSNE().fit_transform(hd)

## Computing local pointwise distortions
spec = [{"id": "snc", "params": {"k": 50}}]
zadu_obj = zadu.ZADU(spec, hd, return_local=True)
global_, local_ = zadu_obj.measure(ld)
l_s = local_[0]["local_steadiness"]
l_c = local_[0]["local_cohesiveness"]

## Visualizing local distortions
fig, ax = plt.subplots(1, 2, figsize=(20, 10))
zaduvis.checkviz(ld, l_s, l_c, ax=ax[0])
zaduvis.reliability_map(ld, l_s, l_c, ax=ax[1])


\end{lstlisting}

\section{Conclusion}

Utilizing distortion measures has so far been time-consuming due to the lack of a well-established implementation.
To address this issue, we present \library, a Python library that allows easy and scalable execution of distortion measures. 
We believe that \library will mitigate the challenges associated with the evaluation of DR embeddings, promoting the design and development of visual analytics applications for high-dimensional data. 

We plan to extend our library into JavaScript, making it compatible with a wider range of existing visualizations \cite{bostock11tvcg} and DR \cite{cutura20vis} toolboxes. Investigating how each distortion measure operates in more detail will also be an interesting direction. We would also like to provide guidelines for utilizing distortion measures.

\acknowledgments{
This work was supported by the National Research Foundation of Korea (NRF) grant funded by the Korea government (MSIT) (No. 2023R1A2C200520911).}

\bibliographystyle{abbrv-doi}

\bibliography{ref}
\end{document}